%% file: main.tex
\pdfoutput=1

\documentclass[11pt]{article}

\usepackage[final]{acl}
\usepackage{arydshln}
\usepackage{xcolor}
\definecolor{ForestGreen}{RGB}{34,139,34}  
\usepackage{enumitem}  
\usepackage{listings}
\usepackage[most]{tcolorbox}
\usepackage{latexsym}
\usepackage{mdframed} 
\usepackage{amsmath}     
\usepackage{booktabs}   
\usepackage{array}       
\usepackage{amssymb}     
\usepackage{latexsym}    
\usepackage{amsfonts}    
\usepackage{algorithm}   
\usepackage{algpseudocode}  
\usepackage{csquotes}    
\usepackage{graphicx}    
\usepackage{stfloats}    
\usepackage{hyperref}    
\usepackage{amsthm}      
\usepackage{multirow}    
\usepackage{pifont}      
\usepackage{siunitx}     
\usepackage{tikz}        
\usetikzlibrary{arrows.meta}  
\usepackage{multirow}

\definecolor{skyblue}{RGB}{0, 102, 204}
\usepackage{amsmath}
\usepackage{booktabs}
\usepackage{array}
\usepackage{amsmath}
\usepackage{amssymb}
\usepackage{times}
\usepackage{latexsym}
\usepackage{times}
\usepackage{latexsym}
\usepackage{amsmath}
\usepackage{amsfonts}
\usepackage{amssymb}
\usepackage{algorithm}
\usepackage{algpseudocode}
\usepackage{csquotes} 
\usepackage{graphicx}
\usepackage{stfloats}
\usepackage{amsmath}
\usepackage{amssymb}
\usepackage{graphicx}
\usepackage{hyperref}
\usepackage{amsthm}
\usepackage{amssymb}
\usepackage{marvosym} 
\usepackage{booktabs}
\usepackage{multirow}
\usepackage{amssymb}
\usepackage{caption}
\usepackage{graphicx}
\usepackage{array}
\usepackage{booktabs}
\usepackage{makecell}
\usepackage{array}
\usepackage{booktabs}
\usepackage{multirow}
\usepackage{graphicx}
\usepackage{xcolor}
\definecolor{ForestGreen}{RGB}{34,139,34} 
\usepackage{colortbl} 
\usepackage{geometry}
\usepackage{caption}
\usepackage{array}

\usepackage{graphicx} 
\usepackage[utf8]{inputenc} 
\usepackage{tcolorbox}
\tcbuselibrary{breakable} 
\usepackage{pifont} 
\usepackage[T1]{fontenc}
\usepackage{microtype}
\usepackage{inconsolata}
\usepackage{graphicx}
\usepackage{booktabs} 
\usepackage{siunitx} 
\usepackage{tikz}
\usetikzlibrary{arrows.meta}
\usepackage{graphicx}
\usepackage{tikz}
\usetikzlibrary{arrows.meta}

\usepackage{amssymb}

\usepackage{inconsolata}
\usepackage{tabulary}

\title{Speculative Reward Model Boosts Decision Making Ability of LLMs Cost-Effectively}

\author{
  Jiawei Gu \\
  Sun Yat-sen University \\
  \texttt{kuvvius@gmail.com} \\
  \And
  Shangsong Liang\textsuperscript{\Letter} \\
  Sun Yat-sen University \\
  \texttt{liangshangsong@gmail.com}
}

\begin{document}
\maketitle

\begingroup
  \renewcommand\thefootnote{\Letter}
  \footnotetext{Corresponding author.}
\endgroup

\begin{abstract}
Effective decision-making in Large Language Models (LLMs) is essential for handling intricate tasks. However, existing approaches prioritize performance but often overlook the balance between effectiveness and computational cost. To address this, we first introduce the 3E Criteria to systematically assess the cost-effectiveness of search strategies, revealing that existing methods often trade significant efficiency for marginal performance gains. To improve LLM decision-making while maintaining efficiency, we propose the Speculative Reward Model (SRM), a plug-and-play framework that seamlessly integrates with existing search strategies. Specifically, SRM employs an external reward assigner to predict optimal actions, reducing reliance on LLMs' internal self-evaluation. And a speculative verification mechanism is used to prune suboptimal choices and guide the search toward more promising steps. We evaluate SRM on several complex decision-making tasks including mathematical reasoning, planning and numerical reasoning in specialized domains. Experimental results show that SRM reduces costs to 1/10 of the original search framework on average while maintaining effectiveness.
\end{abstract}

\input{tex/1_intro}

\input{tex/2_Speculative_Reward_Model}

\input{tex/3_Experiment}

\input{tex/4_Related_Work}

\input{tex/5_Conclusion}

\input{tex/6_Limitation}

\bibliography{custom}

\clearpage
\appendix

\input{tex/7_Appendix}

\end{document}

%% file: tex/1_intro.tex
\section{Introduction}

Large Language Models (LLMs)~\citep{2023gpt4,o12024,deepseek2024r1lite,qwq-32b-preview} have achieved significant progress in natural language processing, excelling in text generation and comprehension~\citep{xu2025towards_reasoning_models_survey}. 
However, their application to complex reasoning and decision-making remains challenging~\citep{deepseekmath, quiet-star}, particularly when solving intricate problems that require structured logical inference rather than pattern-based predictions~\citep{planbench,deepseekmath}.

\input{tables/3e_table}

To address these limitations, early studies introduced prompting strategies to enhance reasoning, such as Chain-of-Thought~\citep{CoT} and 
AlphaZero-Like Tree-Search Method~\citep{wan2024alphazero}, which guide LLMs to generate intermediate reasoning steps to improving inference structure and accuracy. However, these methods rely solely on prompting without external validation or optimization~\cite{song2025prmbench}, limiting their reliability.
Recent approaches employ tree-based search algorithms~\citep{GoT,XoT,agentq,chain-of-table} to explore broader reasoning paths and refine intermediate steps. By systematically evaluating multiple candidates in test time scaling~\cite{tt_scaling}, these methods enhance both the quality and diversity of reasoning, leading to more robust decision-making.

Despite these improvements, they inevitably introduce substantial computational cost. In Table~\ref{tab:3E}, we utilize our proposed \textbf{\textit{3E Criteria}}---\textit{\textbf{E}ffectiveness}, \textit{\textbf{E}fficiency}, and \textit{\textbf{E}xtensibility} to assess the cost incurred during LLM inference. 
\textit{Effectiveness} represents the success rate, \textit{Efficiency} denotes the time and token cost, and \textit{Extensibility} is the adaptability to new tasks. 

The results reveal that existing methods offer limited performance gains at disproportionately high costs. For example, ToT~\citep{ToT}, which employs Depth-First Search (DFS), Breadth-First Search (BFS), provides marginal performance improvements (0-3\%), but incurs a 50-60$\times$ in time cost and a 100-120$\times$ escalation in inference complexity.
 Similarly, RAP~\citep{RAP} leverages Monte Carlo Tree Search (MCTS), yielding a modest performance improvements of 4-5\% at the expense of a 150-300$\times$ increase in inference cost. Additionally, Toolchain*~\citep{toolchain} and reasoning enhanced models like QwQ~\cite{qwq}, constrained by task-specific heuristics, fails to reduce cost effectively and lacks extensibility. \textbf{In this work, we seek to address: }

\vspace{-5pt}
\begin{tcolorbox}[colback=lightgray!10, colframe=black, title={Research Question}]
How to improve the reasoning ability of LLMs while maintaining a balance between effectiveness, efficiency, and extensibility?

\end{tcolorbox}
\vspace{-5pt}

Inspired by studies~\cite{cant_self_correct} emphasizing the need for external validation in decision-making, we propose \textbf{Speculative Reward Models (SRM)}, a plug-and-play framework designed to balance effectiveness and efficiency~\citep{plato}. SRM introduces external rewards to mitigate ineffective decision-making in a speculative manner~\citep{xu2024canSpS,speculativesampling,speculative-decoding}. It consists of two key components: (1) SRM, an independent reward model that assigns scores based on decision consistency and goal alignment. (2) Speculative Verification, a mechanism that ranks candidate steps by evaluating the consistency between internal rewards from LLMs and external rewards from SRM, enabling efficient pruning of suboptimal choices and guiding the search toward more promising states, thereby reducing computational cost. 

We first train SRM on datasets with weak process rewards and then fine-tune it to SRM$^+$ using strong search rewards. This allows us to provide potential success probabilities for specific steps as external reward signals to LLMs during the search phase. Extensive validation has demonstrated that our approach significantly lowers the cost to a fraction of the original search framework's, without sacrificing effectiveness.
In summary, our contributions are as follows:

\textbf{(1) Efficiency}. The SRM framework we proposed dramatically increases efficiency with a notable reduction in cost, requiring only about 1/10 of the original search paradigms.

\textbf{(2) Effectiveness}. There is no sacrifice of effectiveness for SRM; in fact, by integrating reward signals for process supervision, it achieves a up to a 10\% performance improvement over CoT and approximately a 2\% increase compared to using searching algorithms only.

\textbf{(3) Extensibility}\footnote{Refers to whether the method requires retraining to adapt to new problems across different domains.} SRM provides generalizable weak rewards and a universal framework for deriving strong rewards. Fine-tuning with strong rewards transforms SRM into SRM$^+$, enabling domain-specific adaptation without full retraining.

%% file: tables/3e_table.tex
\begin{table}[!t]
\centering
\captionsetup{font=footnotesize}
\caption{\textbf{Speculative Reward Models (SRM)}, a plug-and-play framework designed to balance effectiveness and efficiency. In GSM8K tasks, all paradigms followed the same setting with \textit{GPT-3.5-turbo} and 4-shot learning. The token cost is expressed in `[Prompt Tokens]/ [Completion Tokens]'. "Ext." denotes Extensibility. 
For Toolchain$^*$, which lacks direct execution capability, we estimate cost using identical prompts but exclude running time.}
\label{tab:3E}
\vspace{-0.8em}
\resizebox{\columnwidth}{!}{
\begin{tabular}{l c >{\centering\arraybackslash}p{1cm} >{\centering\arraybackslash}p{1.5cm} c}
\toprule
\multirow{2}{*}[-3ex]{\textbf{Paradigm}}  & \multirow{2}{*}[-3ex]{\textbf{Effectiveness}} & \multicolumn{2}{c}{\textbf{Efficiency}} & \multirow{2}{*}[-3ex]{\textbf{Ext.}} \\
\cmidrule(lr){3-4}
  & & \textbf{Time Cost} & \textbf{Token Cost} & \\
  & Acc.[\%] & Avg.[sec.] & Avg.[K] & \\
\midrule

CoT\citep{CoT}  & 70.1 & 3.2 & 0.7/0.1 & \checkmark \\

\cmidrule(lr){1-5}

DFS\citep{ToT} &  69.9 & 150 & 70.2/5.0 & \checkmark \\
\quad  \textbf{+ SRM}  & 70.5 & 34.7 & 18.6/0.8 & \checkmark \\

\cmidrule(lr){1-5}

BFS\citep{ToT}   & 72.3 & 180 & 85.5/7.1 & \checkmark \\
\quad  \textbf{+ SRM} & 70.1 & 44 & 22.2/1.1 & \checkmark \\

\cmidrule(lr){1-5}

BS\citep{wan2024alphazero}  & 71.4 & 66.4 & 225.4/4.4 & \checkmark \\
\quad  \textbf{+ SRM}  & 72.3 & 44 & 30.8/1.1 & \checkmark \\

\cmidrule(lr){1-5}

MCTS\citep{RAP}   & 74.7 & 122.6 & 105.2/2.5 & \checkmark \\
\quad  \textbf{+ SRM}  & \textbf{80.5} & \textbf{45.2} & \textbf{20.6/0.9} & \checkmark \\

\cmidrule(lr){1-5}

\makecell[l]{Toolchain* \\ \citep{toolchain}} & 78.9 & -- & 40.8/1.9 & $\times$ \\

\bottomrule
\end{tabular}
}
\vspace{-2em}
\end{table}

%% file: tex/2_Speculative_Reward_Model.tex
\section{Problem Formulation}

The decision-making process can be formulated as a Markov Decision Process (MDP)~\cite{mdp}, where the state space $\mathcal{S}$ represents all possible problem states with $s \in \mathcal{S}$, and the action space $\mathcal{A}$ consists of actions $a \in \mathcal{A}$ that transition the state toward a solution. The LLM acts as a generator $\mathcal{G}$, producing candidate actions $\mathcal{G}(a|s, prompt_1)$ and determining state transitions $\mathcal{G}(s'|s,a, prompt_2)$. A reward function $\mathcal{R}(s,a)$ evaluates the effectiveness of actions in progressing toward the goal.

\begin{figure*}[t]
    \centering
    \includegraphics[width=\textwidth]{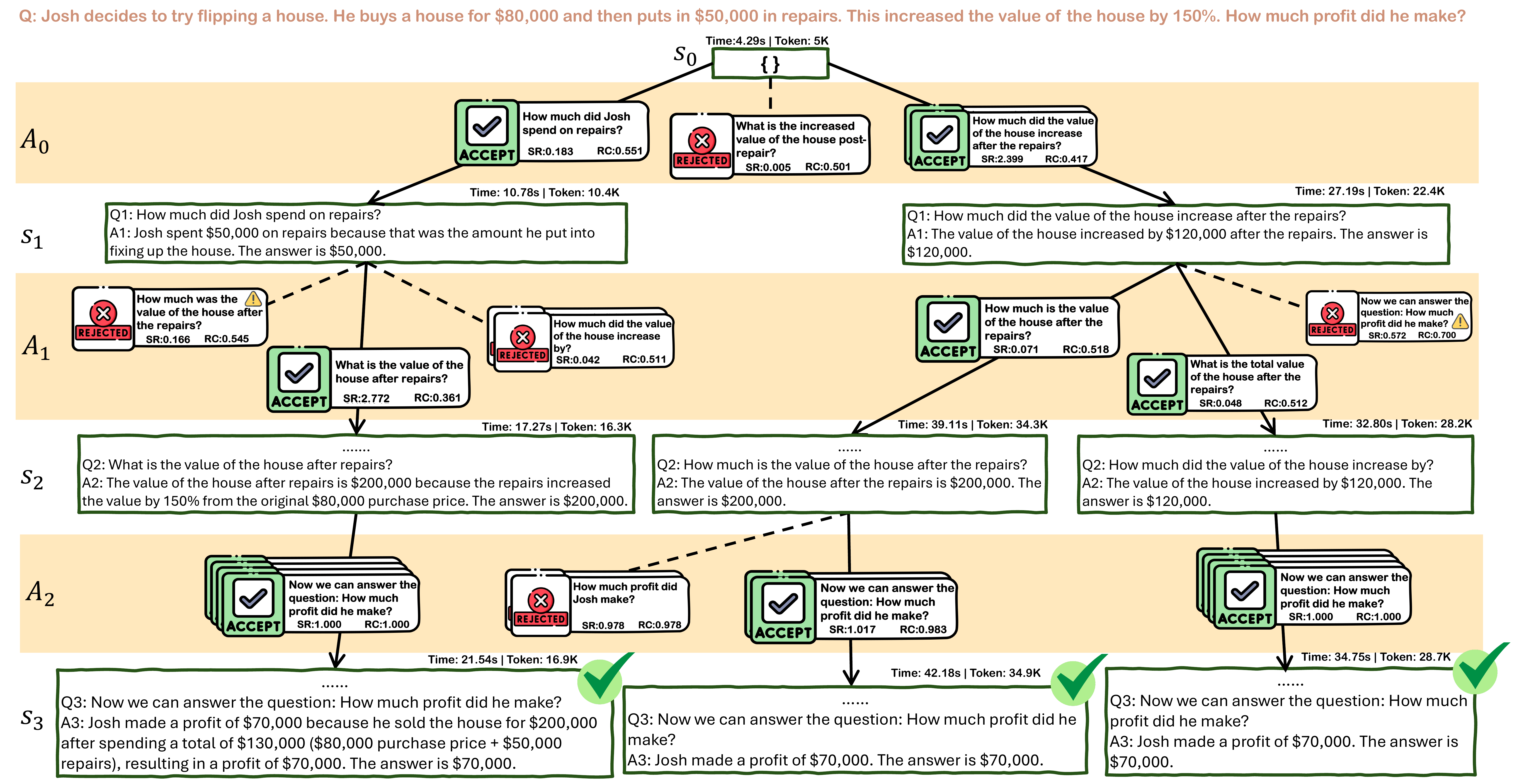}
    \caption{An example in GSM8K ($K=4,N=5$), where our SRM uniquely solves the case correctly across all baselines in 10 tests while achieving the lowest time and token costs. The decision-making process showcases SRM's pruning via Speculative Reward ($SR$), with green actions for acceptance and red for rejection. By $SR$, searching bypasses bad nodes and expands promising ones first. The selection strategy is determined by Reward Consistency ($RC$), prioritizing high-$RC$ actions for earlier development, streamlining the path to the goal. 'Dangerous' sub-questions, characterized by excessively large spans (\includegraphics[height=2ex]{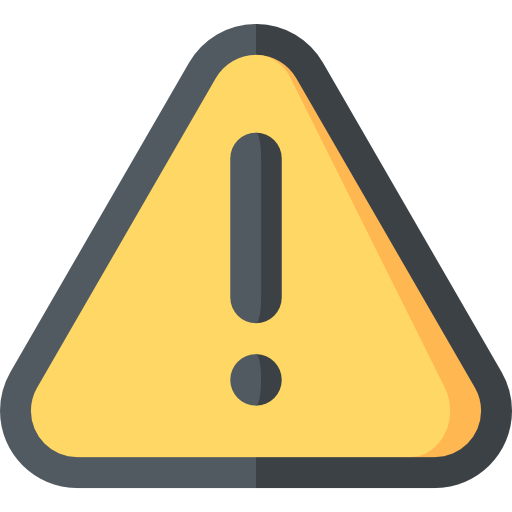}), are pruned efficiently.}
    \label{fig:srm_good_case}
\end{figure*}

Tree-based search paradigms in LLMs decompose complex problems into a sequence of manageable sub-problems, each represented as an action modifying the current state toward the final solution. The search tree $\mathcal{T} = (\mathcal{S},\mathcal{A})$ in Figure~\ref{fig:srm_good_case} represents the decision process, where nodes are states and edges are actions. Starting from an initial state $s_0$, LLM iteratively generates candidate actions $A_n = \{ a_n^{i} \}_{i=1}^K$, assigns rewards $r_{a_n^i} = \mathcal{R} (s_n,a_n^i)$, selects the optimal action $a_n^*$, and transitions to the next state $s_{n+1}$. The search process continues until the goal state $s_g$ is reached, optimizing the cumulative expected reward along the way (see Algorithm~\ref{algo:tree-based_search}).

\section{Method}\label{sec:method}

In this section, we introduce our \textbf{SRM} framework across three key dimensions:  
(1) Speculative Reward (SR) for \textit{\textbf{E}fficiency}, reducing computational cost by pruning less promising search paths;  
(2) Reward Consistency (RC) for \textit{\textbf{E}ffectiveness}, ensuring stable and reliable decision-making by aligning internal and external reward signals;  
(3) SRM$^+$ for \textit{\textbf{E}xtensibility}, enabling adaptation to diverse tasks with minimal retraining.

\paragraph{Speculative Reward for \textit{\textbf{E}fficiency}}

Search strategies typically rely on invoking LLMs to evaluate each state-action pair $(s,a)$, determining the reward $\mathcal{R} (s,a)$. While effective, frequent LLM calls across large search spaces introduce significant inefficiencies. Inspired by Speculative Sampling~\cite{xu2024canSpS,speculativesampling}, which accelerates inference by using a smaller model to \textit{speculate} a larger model’s predictive distribution, we propose the \textbf{SRM} to mimic the LLM as a reward assigner.

Given a pre-order state node $s_n$, and $K$ candidate actions $A_n=\{{a}_{n}^{1}, \ldots, {a}_{n}^{K}\}$ generated from the LLM Generator $\mathcal{G}(\cdot)$, SRM assigns a speculative reward $\mathcal{R}^{\text{SRM}}_\theta (s_n,a_n^i)$ for each action $a_n^i$ as:
\begin{equation}
    \mathcal{R}^{\text{SRM}}_\theta (s_n,a_n^i) = P_{\theta}(a_n^i|s_n,prompt_1),
\end{equation}
where $\theta$ is the parameter of SRM.

By bypassing LLMs for reward assignment, SRM significantly accelerates the search process. To maintain alignment with LLMs priors, following~\citet{speculativesampling}, the reward $ \mathcal{R}^{\text{SRM}}_\theta (s_n,a_n^i)$ for ${a}_{n}^i$ is accepted with probability:
\begin{equation}
\label{eq:accept_prob}
    \min \left( 1, \frac{\bigoplus(P_{\text{LLM}}(a_n^i|s_n, prompt_1))}{\bigoplus(\mathcal{R}^{\text{SRM}}_\theta (s_n,a_n^i))} \right),
\end{equation}
where $\bigoplus(\cdot)$ denotes the normalization operator:

\begin{equation}
   \bigoplus(f(x)) = \frac{f(x)}{\sum_x f(x)}.
\end{equation}
Notably, $P_{\text{LLM}}(a_n^i|s_n, prompt_1)$ is directly obtained from the generation process of $a_n^i$, eliminating additional LLMs queries. Once the action $a_n^i$ is accepted, we update $a_{n}^* \leftarrow {a}_{n}^i$ and transition to the next state $s_{n+1}$ by $\mathcal{G}(s_{n+1}|s_n,a_n^*, prompt_2)$.This process is repeated for ${a}_{n+1}$ until either the goal conditions are met or the search reaches the depth limit. If all actions ${a}_{n}^i (i=1,2,\cdots,K) $ are rejected, we regenerate a new candidate action set $A_{n}'$ from Generator $\mathcal{G}(\cdot)$ and repeat the above process (See Algorithm~\ref{algo:sampling_algorithm}). 

\paragraph{Reward Consistency for \textit{\textbf{E}ffectiveness}} 
Given the speculative property of the ratio in Equation~\ref{eq:accept_prob}, we define it as the Speculative Reward ($SR$), a key metric in our algorithm for pruning. However, assessing absolute performance alone is insufficient, the consistency of reward signals must also be considered. To this end, we propose Reward Consistency ($RC$) as a selection criterion, quantifying the alignment between internal generator rewards and external SRM rewards. It is defined as:
\begin{equation}
RC = \frac{1}{1 + \left| SR - 1 \right|} \in [0, 1]\,.
\end{equation}
An $RC$ value of $1$ indicates complete consistency between internal and external reward signals. Their role within our SRM framework are illustrated in Figure~\ref{fig:srm_good_case}.
Ultimately, the cumulative reward across states (or nodes) is computed by ${R_{\text{accumulated}}} = {{SR}}^{\alpha} \cdot {{RC}}^{(1 - \alpha)}$

 where $\alpha$ is a hyperparameter that balance the significance of ${SR}$ and ${RC}$.

\paragraph{SRM Training and Fine-tuning } The SRM is trained on weak reward labels for each reasoning step—positive, negative, and neutral~(see Appendix~\ref{appendix:weak_reward} for details). Specifically, it is optimized using a cross-entropy loss function to distinguish the more advantageous action among candidates:

\begin{align}
\label{eq:loss_function}
    & \text{loss} (\theta) =  -\frac{1}{\binom{K}{2}} \mathbb{E}_{(s_n,a_n^{i},a_n^{j}) \sim D} \\
& \left[ \log \left( \sigma (\mathcal{R}^{\text{SRM}}_{\theta} (s_n, a_n^{i}) \right.\right. \left.\left. - \mathcal{R}^{\text{SRM}}_{\theta} (s_n, a_n^{j}) ) \right) \right], \nonumber
\end{align}
where \( \mathcal{R}^{\text{SRM}}_{\theta}(s_n, a_n) \) represents the scalar reward assigned by SRM for preorder state \( s_n \) and available action \( a_n \), parameterized by \( \theta \). The model favors actions that lead toward the solution, assigning them higher rewards and the dataset \( D \) contains process-supervised reward or tree-based search reward. This training approach leverages differences in weak rewards to guide SRM in quantifying the intuitive preference for actions that move toward the goal state, thereby enhancing its ability to evaluate the potential success of reasoning steps. Following ~\cite{instructgpt}, all \( \binom{K}{2} \) comparisons from each prior state $s_0$ are processed efficiently as a single batch element to mitigate overfitting.

\paragraph{SRM$^+$ for \textit{Extensibility}} SRM$^+$ is fine-tuned from SRM with same loss described in Equation~\ref{eq:loss_function},  but with a distinct \textit{RewardTuning} dataset. This dataset includes step-level, strong rewards with specific values derived from tree-based search techniques for targeted tasks. Thus, at this stage, SRM$^+$ is more accurate to learn the relative quality of movements through strong labels. The evolution from SRM to SRM$^+$ is illustrated in Figure~\ref{fig:srm_and_srm+}. Besides, further details on the training and fine-tuning methodologies are available in Appendix~\ref{appendix:srm_details}, with data collection for the \textit{RewardTuning} dataset detailed in Appendix~\ref{appendix:rewardtuning}.

%% file: tex/3_Experiment.tex
\section{Experiment}\label{experiment}

In this section, we demonstrate the superiority of the SRM framework~\footnote{Code available at: \url{https://github.com/Kuvvius/Speculative-RM}} in terms of Efficiency, Effectiveness, and Extensibility through comprehensive experiments. 
We evaluate SRM across a diverse range of decision-making scenarios, including mathematical reasoning on GSM8K~\cite{gsm8k}, reasoning and planning in BlocksWorld~\cite{planbench}, and financial numeric reasoning on FinQA~\cite{finqa}. Table~\ref{tab:task-alignment} concisely aligns the three tasks with the decision-making problem framework.

\begin{figure}[t]
  \centering
    \includegraphics[width=0.5\textwidth]{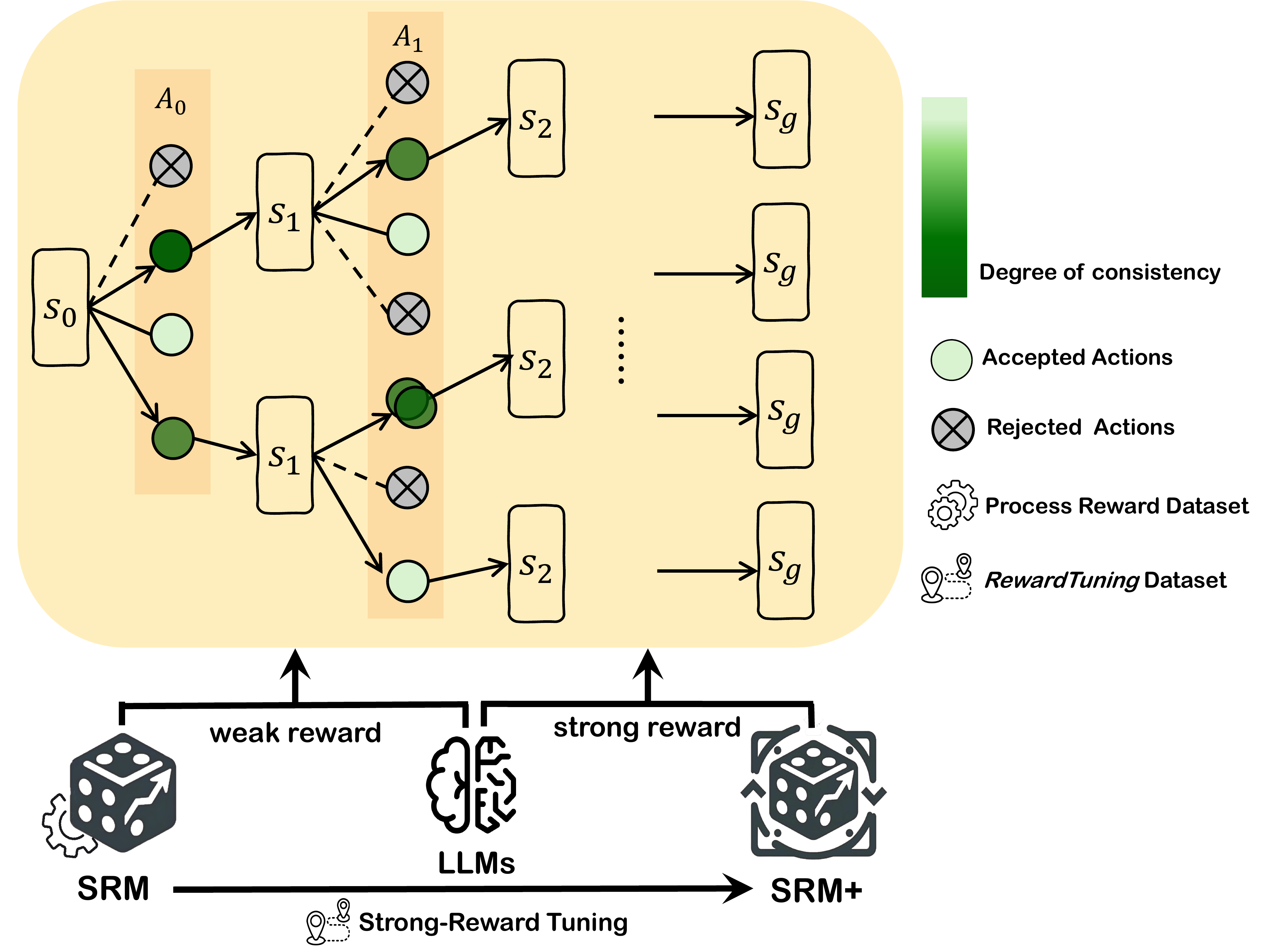}
    \vspace{-2em}
    \caption{SRM was trained using the \textit{PRM800K} dataset, in conjunction with LLMs, to provide weak Speculative Rewards (SR) for each action. Subsequently, SRM$^+$ underwent fine-tuning with the \textit{RewardTuning} dataset, enabling it to generate strong SR for task-specific actions. Various actions are denoted by dots, with the intensity of their green hue indicating the magnitude of the Reward Consistency (RC) on each accepted node. A deeper green signifies a larger RC.}
    \label{fig:srm_and_srm+}
    \vspace{-1.5em}
\end{figure}

\vspace{-0.2em}
\subsection{Experiment Setup}

As shown in Figure~\ref{fig:srm_good_case}, we set $K=4$ (number of candidate actions per step) and $N=5$ (maximum search depth) for all tasks in our experiments.

A detailed discussion of the GSM8K task is presented, while further information on BlocksWorld and FinQA, including their setups and case studies, can be found in Appendix~\ref{appendix:task_details}. Details regarding implementation specifics like SRM configuration, baseline alignment, and our selection of \textit{DeBERTa-v3-large} as the base model are provided in Appendix~\ref{appendix:implementation_details}. Moreover, prompts used in each task are available in Appendix~\ref{appendix:prompt}.

\begin{figure}[t]
  \centering  \includegraphics[width=0.8\columnwidth]{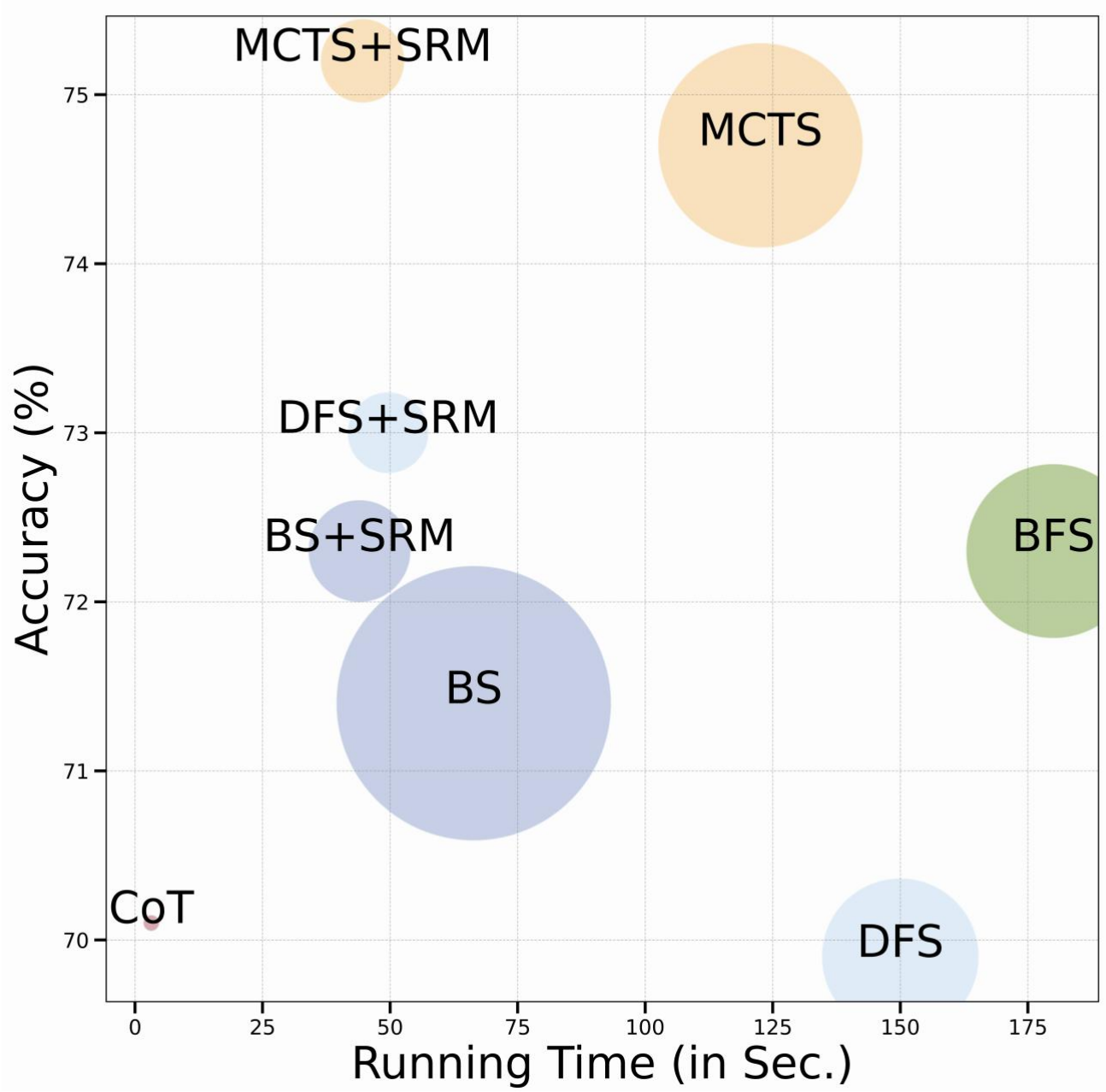}
  \vspace{-0.6em}
    \caption{Comparison of the effectiveness and efficiency of search methods using the plug-and-play SRM framework. The bigger the dot is, the larger the token cost. After applying the SRM framework, it is obvious that the running time of the point representation is reduced~($\leftarrow$), and the accuracy is flat or increased~($\uparrow$).}
    \label{fig:bubbles}
\end{figure}

\subsection{\textit{Effectiveness} and \textit{Efficiency} Analysis}\label{gsm8k-experiment}

\input{tables/gsm8k_result}

To evaluate the impact of SRM on effectiveness and efficiency, we present results on GSM8K from \text{GPT-3.5-turbo} and the LLaMA series~\cite{llama2,grattafiori2024llama} in Table~\ref{tab:3E} and Table~\ref{tab:gsm8k_result}. The results show that SRM significantly reduces both time and token costs by nearly 90\% while maintaining or improving performance (Figure~\ref{fig:bubbles}). Notably, these benefits come without compromising extensibility.

SRM applied to LLaMA-2-70B improves accuracy by 2\% on ToT-DFS and 1\% on RAP-MCTS. When used with GPT-3.5, its cost is only 10\% to 30\% of the original search algorithms. However, results highlight the instability of search paradigms in decision-making tasks. DFS, for example, performs 2\% worse than CoT alone. Integrating DFS with SRM mitigates this decline by pruning weak nodes and expanding stronger ones. The fine-tuned SRM$^+$ further enhances search performance while stabilizing the framework at a lower cost. Additionally, SRM can be fine-tuned using other tree-based search rewards, as discussed in Appendix~\ref{tree-based search reward}. Overall, MCTS+SRM proves to be the most cost-effective approach across GPT-3.5-turbo and the LLaMA series.
Among the evaluated search paradigms, \textbf{MCTS exhibits the highest accuracy yet the highest time cost.} This can be attributed to its more reliable reward system, derived from multiple simulations, rather than the self-evaluation and positional relationship utilized by BFS and DFS. Therefore, in our experiment, we use the MCTS reward in \textit{RewardTuning} as the strong reward label to acquire SRM$^+$. Overall, MCTS+SRM emerges as the most cost-effective approach for decision-making tasks, as demonstrated using GPT-3.5-turbo and the LLaMA series.

\paragraph{Case Study} \textit{SRM mitigates error propagation by prioritizing reliable search paths and pruning error-prone branches.}
Figures~\ref{fig:srm_good_case} and \ref{fig:rap_bad_case} compare MCTS+SRM and MCTS alone, demonstrating how SRM reduces early mistakes that would otherwise propagate through later steps.
SRM prioritizes concise sub-questions with higher $SR$ and $RC$, effectively pruning unreliable branches and guiding search toward more reliable paths. In contrast, MCTS alone struggles to avoid error-prone branches, leading to early mistakes that propagate through later steps.
MCTS relies on fast rewards and LLM self-evaluation, which, while efficient in some cases, often fails to prevent accumulating errors. Without external supervision, minor mistakes can significantly impact tree search algorithms, as LLMs struggle to self-correct. As shown in Figures~\ref{fig:srm_good_case} and \ref{fig:rap_bad_case}, reducing step size and verifying each step prevents errors from compounding, demonstrating SRM’s role in stabilizing search efficiency while maintaining accuracy.

\input{tables/ablation_table}

\paragraph{Ablation Study}
We conduct ablation studies with the MCTS paradigm to evaluate the impact of reject sampling via $SR$ and selection mechanisms via $RC$ (Table~\ref{tab:ablation-study}). The results indicate that both components in SRM's speculative approach contribute to reducing cost while maintaining performance. 
Using only $SR$ for $R_{accumulative}$ significantly lowers cost but also reduces effectiveness. In contrast, relying solely on $RC$ results in a smaller accuracy drop but at the expense of efficiency. Without sampling, cost increases due to the lack of tree pruning, sometimes exceeding the baseline search algorithms. These findings confirm SRM’s effectiveness in optimizing tree-based search performance.

\subsection{\textit{Extensibility} Analysis}
\input{tables/bw_fin_result}
Table~\ref{tab:bw_and_finqa_result} highlights SRM’s adaptability across decision-making tasks. In Blocksworld (BW), CoT with LLaMA-2-70B struggles with planning, while MCTS improves decisions at high computational cost. SRM reduces inference by 7\% while maintaining accuracy, and SRM$^+$ further enhances performance via \textit{RewardTuning} (See Appendix~\ref{appendix:rewardtuning}).  

Beyond planning, SRM seamlessly transfers to FinQA, 
improving accuracy by 5\% with minimal retraining, while SRM$^+$ achieves an 8\% gain. Notably, SRM$^+$ enables GPT-3.5 to match GPT-4 in efficiency, demonstrating its ability to optimize LLMs across domains. By integrating speculative verification and fine-tuning with task-specific rewards, SRM ensures efficient, cost-effective adaptation to new tasks.

%% file: tables/gsm8k_result.tex
\begin{table*}[ht]
\centering
\setlength{\extrarowheight}{1pt}
\setlength{\tabcolsep}{4pt}
\caption{The result we tested 10 times on GSM8K and put on the average accuracy and cost. The values of total running time and total token cost are represented as multiples of the CoT row's value.}
\vspace{-0.8em}
\resizebox{1\textwidth}{!}{%
\begin{tabular}{@{}lccc|ccc|ccc@{}}
\toprule
\multirow{2}{*}{Method} & \multicolumn{3}{c|}{LLaMA-2-70B}  & \multicolumn{3}{c|}{LLaMA-33B}  & \multicolumn{3}{c}{LLaMA-2-13B} \\ 
\cmidrule(rl){2-4} \cmidrule(rl){5-7} \cmidrule(rl){8-10}
& Effe. [Acc.] & Time [×CoT] & Token [×CoT]  
& Effe. [Acc.] & Time [×CoT] & Token [×CoT]  
& Effe. [Acc.] & Time [×CoT] & Token [×CoT] \\ 
\hline
CoT  & 0.54 & 1.0  & 1.0   & 0.29 & 1.0  & 1.0   & 0.20 & 1.0  & 1.0  \\
\hline
DFS  & 0.52 & 28.4 & 1727.2 & 0.25 & 19.4 & 610.9 & 0.19 & 350.7 & 1306.8 \\
\textbf{+ SRM}  
& 0.54~\textbf{\textcolor{ForestGreen}{($\uparrow$)}} & 4.2 & 233.3 
& 0.26~\textbf{\textcolor{ForestGreen}{($\uparrow$)}} & 2.9 & 32.0 
& 0.20~\textbf{\textcolor{ForestGreen}{($\uparrow$)}} & 43.9 & 64.6 \\
\textbf{+ SRM$^+$}  
& 0.55~\textbf{\textcolor{ForestGreen}{($\uparrow$)}} & 4.2 & 241.2 
& 0.28~\textbf{\textcolor{ForestGreen}{($\uparrow$)}} & 2.9 & 32.4 
& 0.24~\textbf{\textcolor{ForestGreen}{($\uparrow$)}} & 42.0 & 69.5 \\
\hline
BFS  & 0.58 & 36.3 & 1133.7 & 0.38 & 37.8 & 237.8 & 0.23 & 368.5 & 661.5 \\
\textbf{+  SRM}  
& 0.55 & 3.4 & 133.9  
& 0.35 & 2.1 & 41.5  
& 0.23 & 19.5 & 48.5  \\
\textbf{+ SRM$^+$}  
& 0.59~\textbf{\textcolor{ForestGreen}{($\uparrow$)}} & 3.4 & 123.4 
& 0.38 & 2.2 & 42.2 
& 0.26~\textbf{\textcolor{ForestGreen}{($\uparrow$)}} & 19.2 & 42.2 \\
\hline
MCTS  & 0.61 & 1145 & 295.1 & 0.49 & 74.6 & 108.1 & 0.30 & 61.2 & 180.7 \\
\textbf{+ SRM}  
& 0.62~\textbf{\textcolor{ForestGreen}{($\uparrow$)}} & 8.0 & 66.7 
& 0.49 & 2.2 & 19.9  
& 0.27 & 15.3 & 33.0  \\
\textbf{+ SRM$^+$}  
& 0.64~\textbf{\textcolor{ForestGreen}{($\uparrow$)}} & 8.0 & 63.4  
& 0.51~\textbf{\textcolor{ForestGreen}{($\uparrow$)}} & 2.3 & 20.7  
& 0.29 & 15.3 & 31.8 \\ 
\bottomrule
\end{tabular}%
}
\label{tab:gsm8k_result}
\vspace{-1em}
\end{table*}

%% file: tables/ablation_table.tex
\begin{table}[tb]
\centering
\setlength{\extrarowheight}{1pt} 
\caption{The baseline is MCTS. Sampling refers to the rejection sampling strategy outlined in Section~\ref{sec:method}, absent which there is no pruning. Consistent with earlier sections, token costs are denoted as [Prompt Tokens]/[Completion Tokens].}
\resizebox*{!}{0.47\columnwidth}{
\begin{tabular}{llll@{}} 
\toprule
\multicolumn{1}{c}{{\multirow{2}{*}[-1ex]{\textbf{Method}}}} & \multicolumn{1}{c}{{\multirow{2}{*}[-1ex]{\textbf{Effectiveness}}}} &
\multicolumn{2}{c}{\textbf{Efficiency}} \\
\cline{3-4}
 &  & \textbf{Time Cost} & \textbf{Token Cost}   \\
 & Acc.[\%] & Avg.[Sec.] & Avg.[K] \\
\midrule
MCTS & 74.7 & 122.6 & 105.2/2.5  \\
  + $SR$ + sampling & $70.2_{\downarrow4.5\%}$ & 28.3 & 16.3/0.4\\
  + $RC$ + sampling & $71.4_{\downarrow3.3\%}$ & 96.5 & 53.2/1.5  \\
  + ${{SR}}^{\alpha} \cdot {{RC}}^{(1 - \alpha)}$ + sampling & $80.5_{\uparrow5.8\%}$ & 45.2 & 20.6/0.9  \\
  + $SR$ no sampling & $78.4_{\uparrow3.7\%}$ & 105.1 & 70.8/2.1  \\
  + $RC$ no sampling & $73.3_{\downarrow1.4\%}$ & 143.2 & 98.1/2.7  \\
  + ${{SR}}^{\alpha} \cdot {{RC}}^{(1 - \alpha)}$ no sampling & $75.1_{\uparrow0.4\%}$ & 58.8 & 34.7/0.9  \\
\bottomrule
\end{tabular}}
\label{tab:ablation-study}
\vspace{-1.5em}
\end{table}

%% file: tables/bw_fin_result.tex
\begin{table}[ht]
\caption{Result of Blocksworld~(LLaMA-2-70B) and FinQA~(GPT-3.5 and GPT-4).}
\label{tab:bw_and_finqa_result}
\resizebox*{!}{0.65\columnwidth}{
\centering
\small
\resizebox*{!}{0.50\columnwidth}{
\begin{tabular}{llccc}
\toprule
\bf Mode & \bf Method & \bf Eff. & \bf Time & \bf Token \\
\midrule
\multirow{4}{*}{\shortstack{BW(Easy)}}    
& CoT                 & 0.08  & 1.0x   & 3.8   \\
& MCTS                & 0.66  & 560.9x & 366.0 \\
& \textbf{MCTS + SRM} & 0.66  & 54.4x  & 40.1  \\
& \textbf{MCTS + SRM$^+$} & 0.68 & 58.3x & 47.0  \\
\midrule
\multirow{4}{*}{\shortstack{BW(Hard)}}    
& CoT                 & 0.05  & 1.0x   & 3.8   \\
& MCTS                & 0.51  & 709.5x & 416.7 \\
& \textbf{MCTS + SRM} & 0.49  & 54.8x  & 34.2  \\
& \textbf{MCTS + SRM$^+$} & 0.54 & 69.9x & 45.5  \\
\midrule
\multirow{4}{*}{\shortstack{FinQA\\(GPT3.5)}} 
& CoT                 & 0.49  & 4.5    & 3.4   \\
& MCTS                & 0.60  & 160.6  & 200   \\
& \textbf{MCTS + SRM} & 0.65  & 51.9   & 54.2  \\
& \textbf{MCTS + SRM$^+$} & 0.68 & 52.1  & 53.7  \\
\midrule
\multirow{1}{*}{\shortstack{FinQA (GPT-4)}} 
& CoT                 & 0.70  & 4.9    & 3.5   \\
\bottomrule
\end{tabular}}}
\end{table}


%% file: tex/4_Related_Work.tex
\section{More Discussion}

\paragraph{Diversity and randomness bring stable improvement.}
The methods related to Decision-making agents would have unstable issues and strongly depend on the general ability of the base model. 
During the reasoning process, MCTS introduces a degree of randomness in generating the final results. This randomness, combined with the diversity at intermediate nodes, allows for stable optimization of the sampling outcomes from language models. Consequently, MCTS consistently demonstrates superior performance compared to other search methods.

\paragraph{External signals can effectively supervise the generation process of the content.}
When a decision-making agent engages in complex reasoning and problem-solving, it heavily relies on the generative capabilities of the language model. However, using only self-evaluation methods often fails to provide stable and reliable judgments, making effective process supervision difficult. In such cases, introducing an external verifier for process supervision proves to be effective. The verifier can provide feedback on the quality of the model's current outputs and offer guidance, which helps improve performance.

By leveraging diversity (note that the ``diversity'' here differs from ``diversity'' in the field of information retrieval~\cite{liang2017search,liang2019collaborative}) and randomness, the use of effective external signals for proper guidance can help avoid the high costs associated with repetitive exploration in the search space. Specifically, the verification signals provided by our proposed SRM in domain-specific problems, combined with search methods that allow for \textit{\textbf{sufficient exploration and randomness}}, can achieve cost-effective performance improvements.

\paragraph{Why a relatively small model can help large base model?}\label{weak-to-strong analysis}

Our reward model underwent training that supervised the decision-making process, but it's significantly smaller compared to the generative language models it supports. The feasibility of using a smaller-scale reward model to effectively assist a much larger, more powerful model lies in our acknowledgment of the errors inherent in the weak labels provided by the Supervised Reward Model (SRM). However, within our framework, we do not intend for the more robust model to learn or replicate these errors. Instead, our aim is to guide it toward understanding the intentions behind the supervision (i.e., signals of external oversight), not the inaccuracies themselves. We maintain the assumption that the larger, base model \textit{\textbf{inherently possesses}} all necessary reasoning and decision-making capabilities but might not currently exhibit them due to limitations in the decision-making context. Under the \textit{\textbf{guidance}} of a ``weaker'' model, it becomes possible to activate this latent knowledge and adjust the base model towards a direction of self-reward, thereby enhancing its performance and decision-making processes in alignment with the supervisors' intentions.

\section{Related Work}

\subsection{Decision-Making Agents}
LLM-based decision-making agents, such as XoT~\cite{XoT}, and Quiet-STaR~\cite{quiet-star} generate structured actions using formal languages like PDDL or API calls. These models rely on binary or scalar feedback for policy optimization, differing from human decision-making~\cite{agent-as-a-judge}. Memory-enhanced methods~\cite{reflexion2023,toolchain} treat LLMs as autonomous agents, but reward interpretation remains a challenge~\cite{song2025prmbench}. Our SRM addresses these limitations with a structured, cost-effective decision-making approach.

\subsection{Tree-Based Search Algorithms}
Tree-based search, including DFS, BFS, and MCTS, plays a key role in LLM-driven decision-making~\cite{tt_scaling}. DFS and BFS explore solutions systematically, while MCTS improves decision quality via random sampling. However, methods like ToT~\cite{ToT}, RAP~\cite{RAP} and AlphaZero-Like Tree-Search Method~\citep{wan2024alphazero}  incur high inference costs due to frequent LLM calls.

\vspace{-0.1em}
\subsection{Speculative Sampling}
Speculative sampling~\cite{xu2024canSpS,speculativesampling,speculative-decoding} speeds up LLM inference by drafting candidate tokens and verifying them with a target model, reducing latency while maintaining quality. Inspired by this, SRM applies speculative verification to decision-making, using rejection sampling to prune search paths, minimize redundancy, and improve efficiency.

%% file: tex/5_Conclusion.tex
\section{Conclusion}

We propose the Speculative Reward Model (SRM), a cost-effective framework that enhances LLM decision-making by speculating on potential rewards. SRM reduces ineffective decisions through Speculative-Verification, efficiently ranking steps by given scores. Our contributions include significant cost reductions, a 10\% performance improvement over CoT, a 2\% increase over search-based algorithms, and broad applicability. Additionally, we introduce \textit{RewardTuning}, a dataset for fine-tuning the reward model on three tasks. As to future work, we intend to extend our model for other tasks ~\cite{xian2025molrag,wiki_table_question}.

%% file: tex/6_Limitation.tex
\section*{Limitations}
\paragraph{Dependency on External Models} SRM needs to be fine-tuned with task reward data to improve the corresponding performance on the specific task.
relies on external reward models, which might introduce additional complexity and potential inaccuracies if the external models are not well-calibrated or if they fail to capture the nuances of the specific tasks.

\paragraph{Scalability Challenges} While SRM reduces costs and improves efficiency, it is itself a relatively small model with only about 500M parameters. This limited capacity can pose challenges when scaling to more complex tasks or larger datasets, potentially hindering its ability to generalize effectively.

\section*{Acknowledgments}

This work has been under development for an extended period and has benefited enormously from ongoing refinement and feedback. We are profoundly grateful to Guanzheng Chen for his invaluable guidance, insightful discussions, and unwavering support at every stage of this project. We also wish to thank Jiahao Song for generously providing the critical resources and infrastructure, especially during the early phases, that made our implementation possible. Their contributions have been instrumental in shaping and advancing this research.

%% file: tex/7_Appendix.tex
\appendix

\section{Implementation Details}\label{appendix:implementation_details}
\begin{figure*}[t]
    \centering
    \includegraphics[width=1\textwidth]{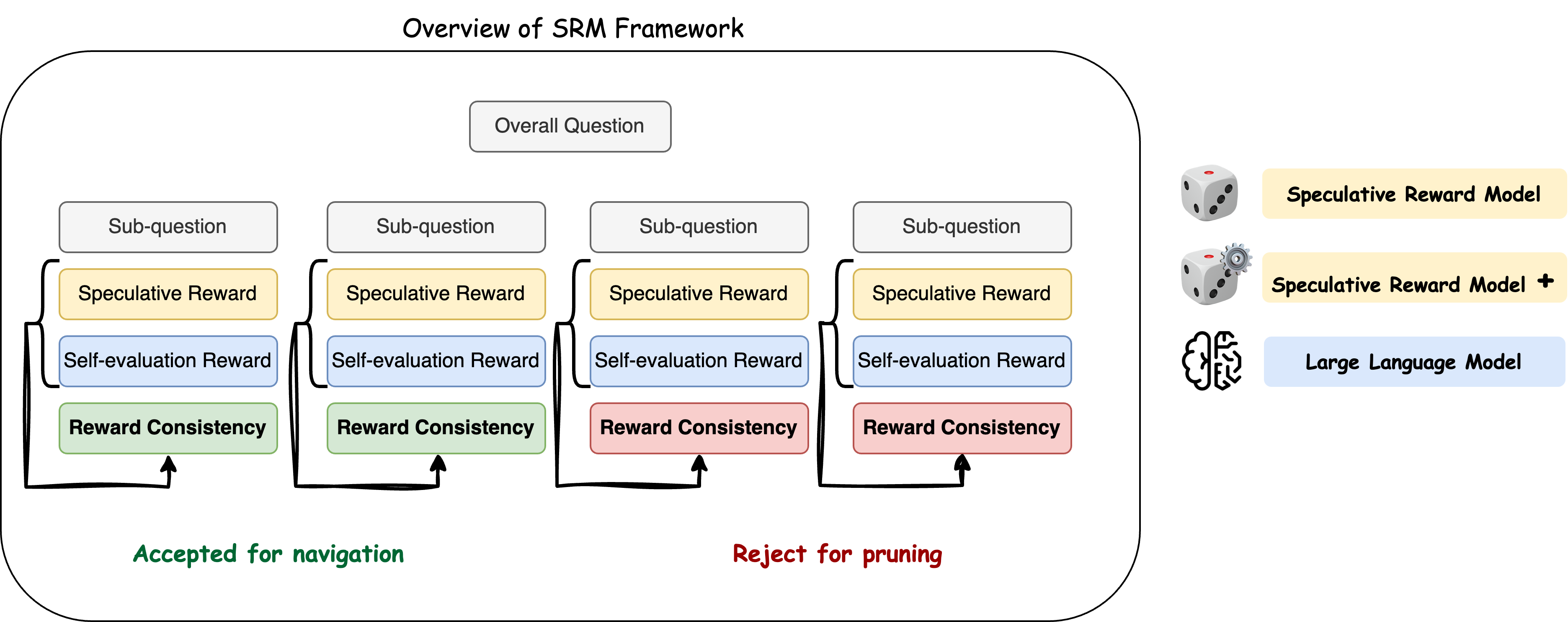} 
    \caption{Example of an efficient selection process}
    \label{fig:overview_working_process}
\end{figure*}

To better illustrate the Decision-making process with SRM, we provide pseudo-code in Algorithm~\ref{algo:sampling_algorithm} and a selection process (including rejection for pruning and acceptance sorting for efficient navigation) as shown in the Figure~\ref{fig:overview_working_process}.

\begin{algorithm}[!t]
\caption{Decision-making process with SRM}\label{algo:sampling_algorithm}
\begin{algorithmic}[1]
\State Given candidate $K$ actions, and depth limit of tree $N$.
\State Given Large Language Model $G(\cdot)$ as generator, and Speculative Reward Model $R(\cdot)$, action-prompt $prompt_1$ and state-prompt $prompt_2$ with few-shot examples, intial state $s_0 = \emptyset$ 

\State Initialise $n \gets 0$.
\While{$n < N$}
    \For{$t = 1 : K$}
        \State Generate candidate actions auto-repressively $ {a_n^t} \sim G(a|s_n,prompt_1)$
    \EndFor
    \State Compute speculative rewards of $K$ candidate actions respectively 
    $ {a_n^t} \sim R(a|s_n,prompt_2)$
    \State $R({a}_n^1 | s_n),\ldots,R(\tilde{a}_n^K | s_n)$

    \For{$t = 1 : K$}
        \State Sample $\epsilon \sim U[0,1]$ from a uniform distribution.
        \If{$\epsilon < \min \left( 1, \frac{\bigoplus(\text{Prob}(a_n^i))}{\bigoplus(\text{Reward}(a_n^i))} \right))$}
            \State Set $a_{n} \gets {a}_n^i$ and $n \gets n + 1$.
        \Else
            \State Continue
        \EndIf
    \EndFor
\EndWhile
\end{algorithmic}
\end{algorithm}

\begin{figure*}[t]
  \centering
    \includegraphics[width=1.5\columnwidth]{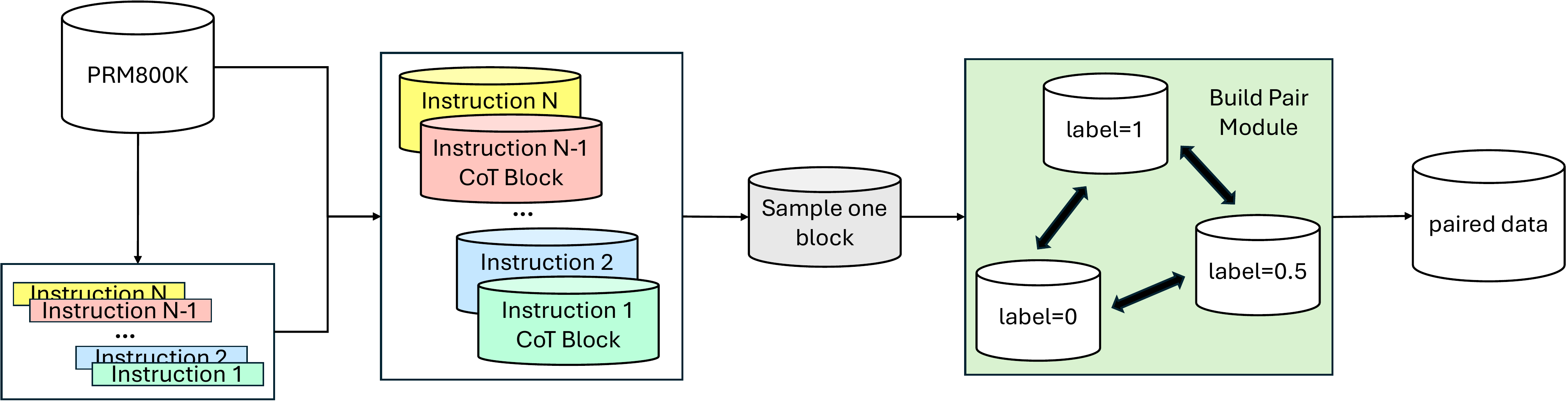}
    \caption{The process of building our weak reward dataset from \textit{PRM800K} dataset, which SRM was trained on. The data samples of state and action pairs can be found in Appendix~\ref{appendix:weak_reward}.}
    \label{fig:weak_reward_builder}
\end{figure*}

\begin{figure*}[t]
    \centering
    \includegraphics[width=0.9\textwidth]{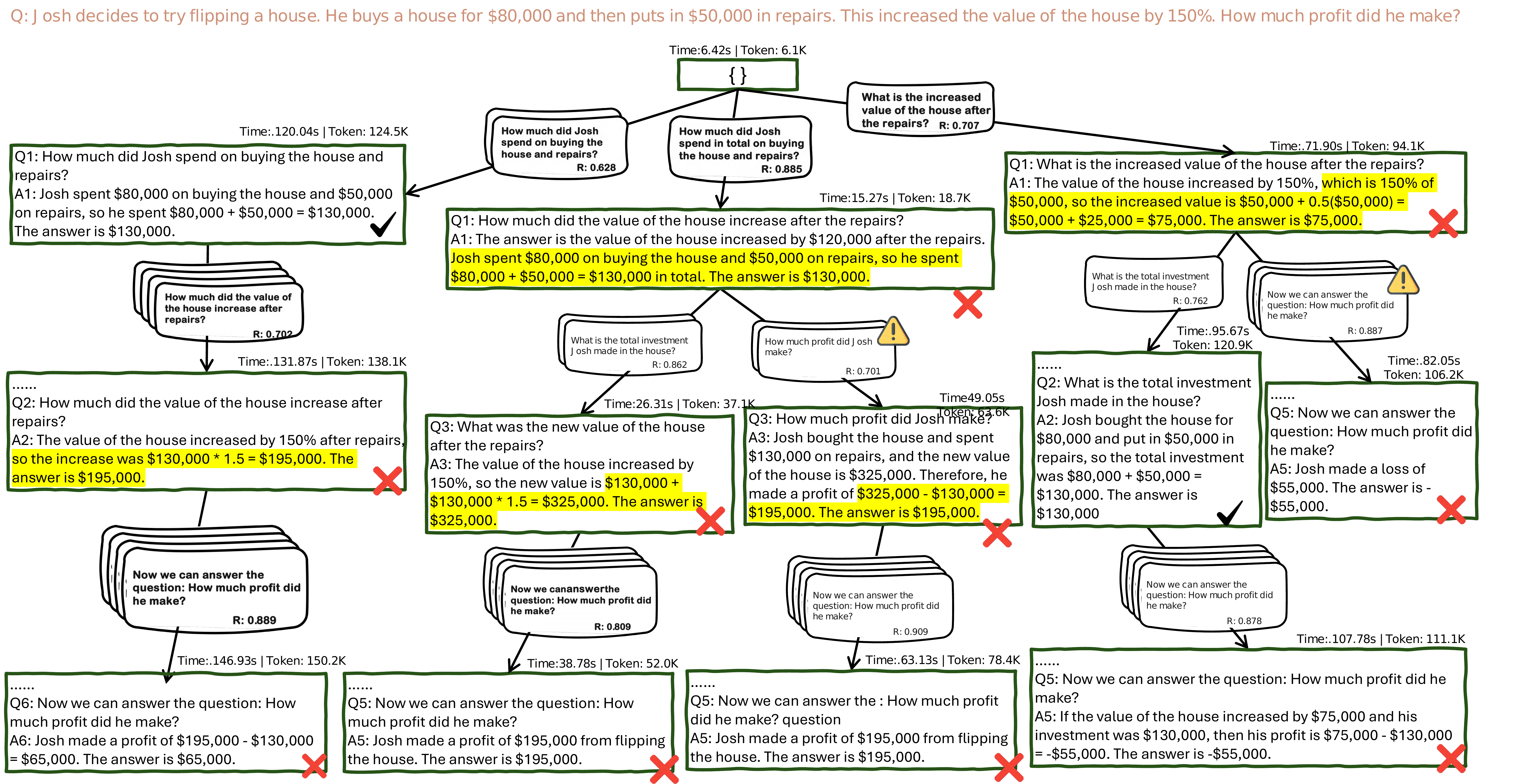} 
    \caption{Bad Case of MCTS Decision-making}
    \label{fig:rap_bad_case}
\end{figure*}

\subsection{LLM Configuration}\label{appendix:srm_details}

In order to align the existing experimental results, we opted for the GPT-3.5-turbo (a previous version) as the engine in constructing the LLM-based agent framework. We configured the solution generation to have a maximum length of 512, with a temperature setting of 0.8, as detailed in Section \ref{experiment}. In the case of LLaMA-2 experiments, we similarly set the maximum solution length at 512 and the temperature at 0.8. The experiments were conducted using 8 NVIDIA Tesla V100 32GB GPUs to facilitate the inference process for both the LLaMA-2 7B and 13B models.

To maintain consistency with the established search algorithms, we adjusted weights as the same as them.

\subsection{SRM Training and Fine-tuning Details}

SRM was trained on DeBERTa-v3-large with sentence pairs with weak labels~\ref{fig:prm800k} to obtain SRM, and fine-tuned by strong labels~\ref{fig:strong_reward_data_sample} evolving into SRM+. As the loss function in Equation~\ref{eq:loss_function}, we train SRM to learn the differences in text with different labels through comparison. Finally, with the input pairs with same state sentence, SRM can give the predicted reward labels, which show relatively good or bad. The dataset we built in our work will be fully released upon acceptance. In the \ref{appendix:weak_reward} and \ref{appendix:rewardtuning}, we provide further clarification and explanations through data samples.

\subsubsection{Process Reward Dataset}\label{appendix:weak_reward}
\begin{figure*}[h]
    \centering
    \includegraphics[width=0.9\textwidth]{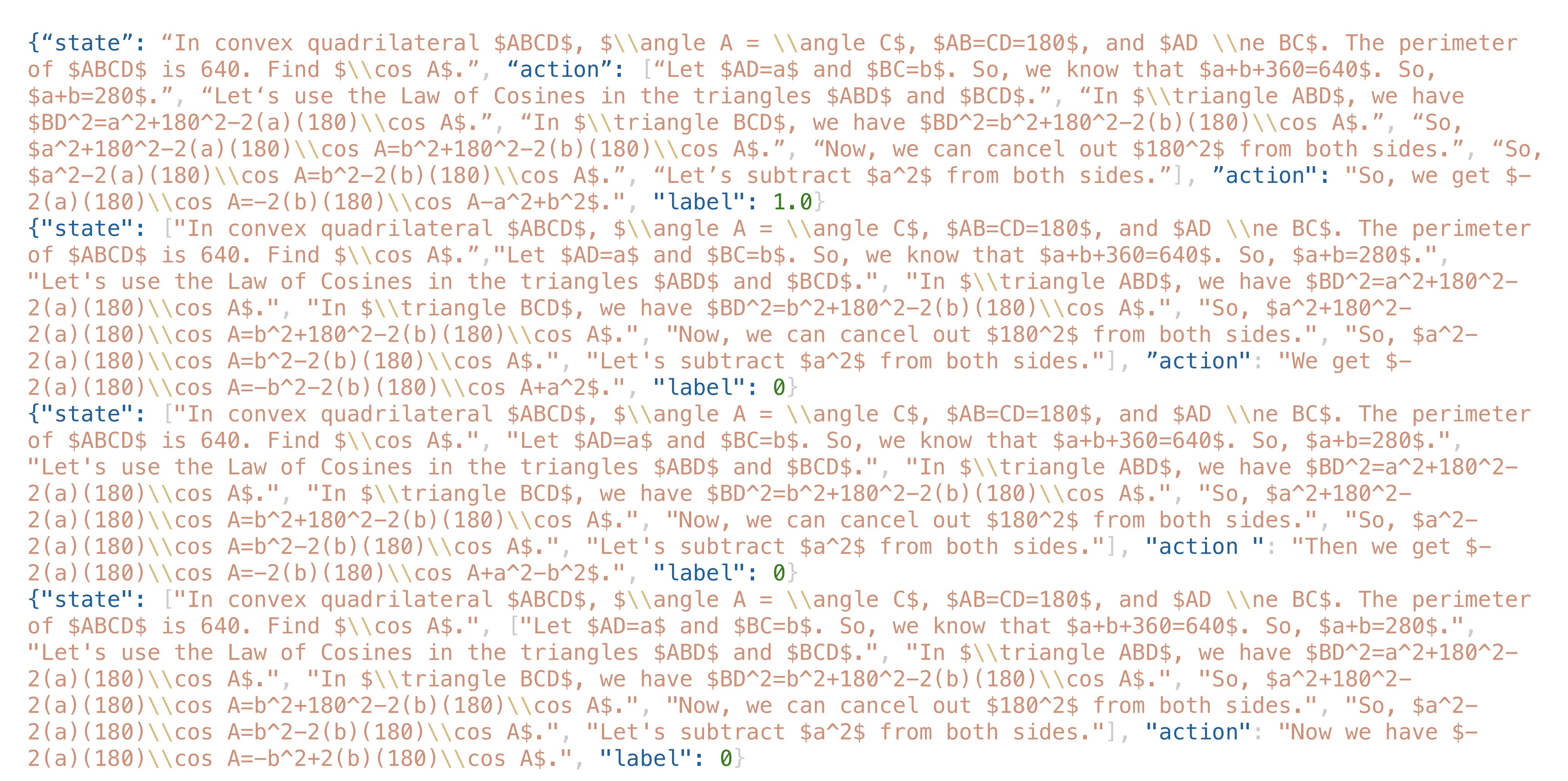} 
    \caption{The process of generating weak reward data pairs. As the example showed, we process the data from prm800k into state and action pairs with labels }
    \label{fig:prm800k}
\end{figure*}

The original training data has 1,055,517 pieces of data and 10,833 instructions (i.e. questions). After processing, there are 3,150,704 pairs. The generating process and data examples are shown in the Figure~\ref{fig:prm800k}.

\subsubsection{\textit{RewardTuning} Dataset}\label{appendix:rewardtuning}
\begin{figure}[h]
    \centering
    \includegraphics[width=\columnwidth]{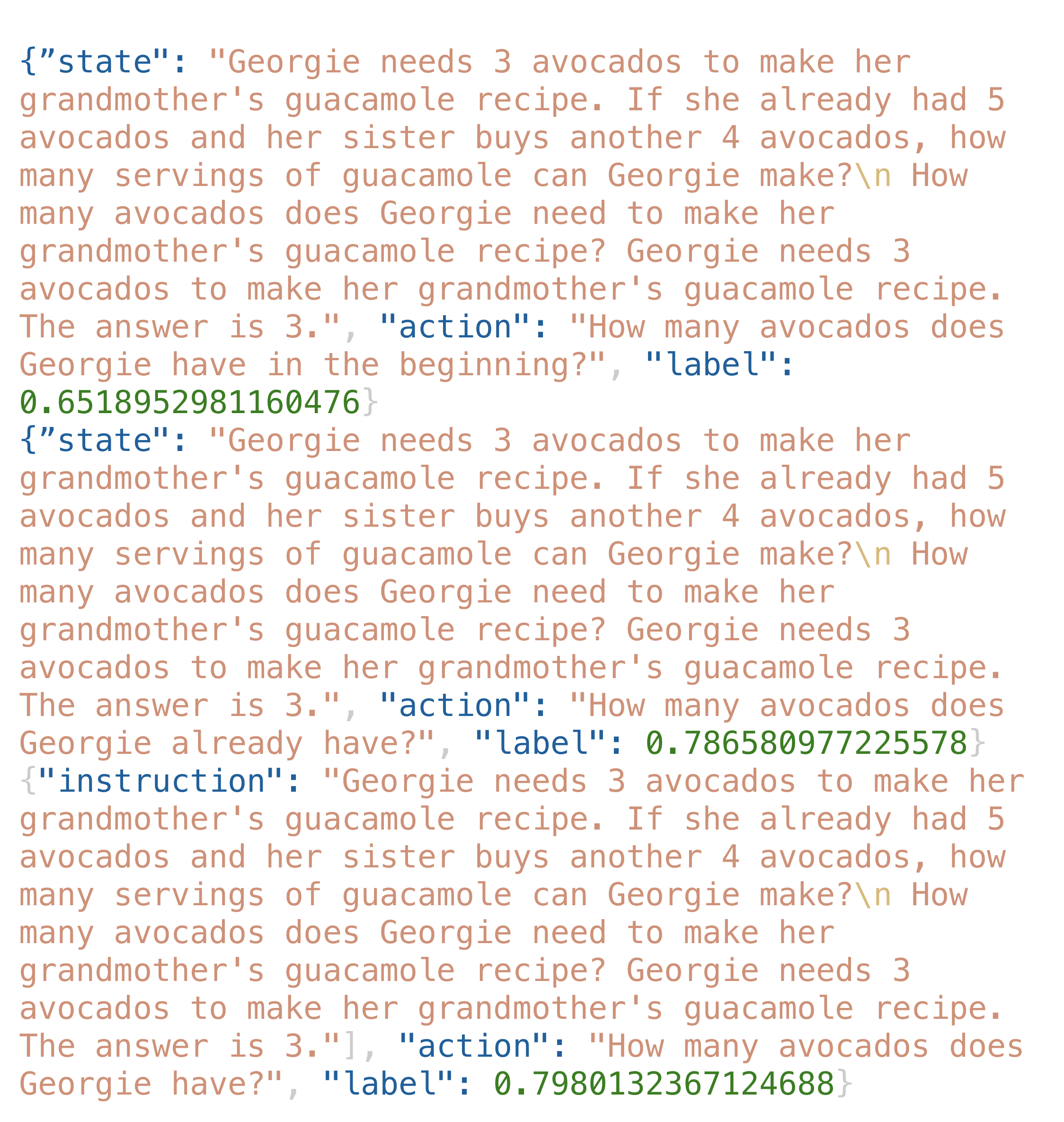} 
    \caption{The process of generating strong reward data pairs.}
    \label{fig:strong_reward_data_sample}
\end{figure}

We use the existing searching method to acquire the strong reward label for each step of sub-question or each state for blocks as shown in Figure~\ref{fig:strong_reward_data_sample}. The form of reward is an exact value. We build all reward data for the training data set of the three task, and finally use 10\% them to fine-tune our SRM$^+$. The generating process and data examples are shown in the Figure~\ref{fig:strong_reward_data_sample}.

\section{More Analysis}

Recent results on GSM8K indicate that while some methods achieve relatively high accuracy, their cost-efficiency remains a major concern. For example, QwQ~\cite{qwq} achieves a very high accuracy of 93.9\%. However, its time cost is not reported (denoted as “-”), and its token cost (0.7/1.6) is only slightly improved relative to baseline methods. Moreover, QwQ is marked with “$\times$” under Extensibility, which means that despite its high performance, its applicability to new tasks is limited due to the reliance on task-specific heuristics (reasoning tasks only).

The results reveal that existing methods offer limited performance gains at disproportionately high costs. For instance, ToT~\cite{ToT}, which employs Depth-First Search (DFS) and Breadth-First Search (BFS), provides only marginal improvements (0–3\%) yet incurs a 50–60$\times$ increase in time cost and a 100–120$\times$ escalation in inference complexity. Similarly, RAP~\cite{RAP} uses Monte Carlo Tree Search (MCTS) to yield a modest performance improvement of 4–5\%, but at the expense of a 150–300$\times$ increase in inference cost. Additionally, while models like Toolchain*~\cite{toolchain} and reasoning-enhanced models like QwQ~\cite{qwq} can achieve high accuracy, they are constrained by task-specific heuristics, fail to reduce cost effectively, and suffer from poor extensibility.

Table~\ref{tab:3E} summarizes the performance (Effectiveness), efficiency (Time and Token Cost) and extensibility of various paradigms in GSM8K tasks under the same setting with \textit{GPT-3.5-turbo} and 4-shot learning. It is evident that despite high effectiveness, models such as QwQ, Toolchain*, and even some search-based paradigms require significant computational resources, whereas methods incorporating Speculative Reward Models (SRM) can offer a better trade-off between performance and efficiency.

\section{Task details}\label{appendix:task_details}

\input{tables/task_illust}

\paragraph{Task Setup}
We evaluate SRM framework with the MCTS search paradigm in Blocksworld benchmark~\cite{planbench}, where the aim is to examine the framework's efficacy in guiding an agent through a sequence of actions to reorganize blocks into specified configurations. In our research, we draw from the Blocksworld dataset as outlined by \cite{planbench}, organizing the test cases by the least number of actions they necessitate for a solution and giving four test case to prompt, as same as \cite{RAP}, which detailed in 
The plan generation task involves creating a sequence of actions to meet the goal, which showcases decision-making skills at each step of the planning process.

\paragraph{BW Result on Step-level}

\input{tables/bw_steps_result}

Importantly, the set of possible actions is finite and determinable through predefined rules rather than requiring generation by an LLM. The action space is dynamically generated, considering both domain-specific constraints and the current orientation of the blocks. For state transitions, the framework consults a Large Language Model (LLM) to forecast the impacts of actions on the blocks' states, updating the current state to reflect new conditions and eliminate outdated ones. The LLM, in conjunction with the SRM, generates Successor Representations ($SR$) and Reward Contexts ($RC$) for potential actions, which then inform the state transition function. The process concludes once the goal state is realized or when the search hits the predetermined depth limit.

\section{Tree-based search Reward}\label{tree-based search reward}

\begin{algorithm}[!t]
\caption{Tree-based Search in LLMs.} \label{algo:tree-based_search}
\begin{algorithmic}[1]
\State \textbf{Input}: $s_0$: input; $G$: large language model; $M$: the maximum exploring steps; $T$: the dynamic decision tree for search; $\mathcal{R}(s_n,a_n^k)$: function to return specific reward
\State \textbf{Initialize} $T = \{S, A\}$; $S \leftarrow s_0$; $A \leftarrow \emptyset$.
\For{$t = 1$ to $N$}
    \State $A_n = \{a^{(i)}\}_{i=1}^{k} \leftarrow G(s_n)$ \Comment{ Invoking}
    \State $a_n^* \leftarrow \arg\max_{a_n \in A_n} \mathcal{R}(s_n,a_n)$ 
    \State Add $a_n$ as the edge of $s_n$.
    \State $s_{n+1} \leftarrow G(s_n,a_n^*)$
    \State Update $s_{n+1}$ as a node of $T$. \Comment{Invoking}
\EndFor
\State \textbf{Output}: The goal state $s_g$ including reasoning steps and answer.
\end{algorithmic}
\end{algorithm}

Rewards are acquired by tree-based search algorithms, different from common reward for language model~\cite{rewardlm,reflexion2023}. 
And all the search methods employed are unsupervised, yet they vary in the balance they strike between exploration and efficient selection.

We would like to detail three kinds of reward designs with the order of decreasing exploration. Besides, we leave the more reward settings
corresponding to the algorithms in the future work. Generally, tree-based search algorithms could own their corresponding reward configure, showing the flexibility of our framework.

\subsection{Priority Reward}
This type of reward are designed for the search with certain priority. Taking DFS for an example, it begins with "root" state $s_0$ and then iteratively choose the first candidate action $a_n^1$ while there are $K$ candidate action nodes. Until it reached the depth limit or the goal state $s_g$ containing the final correct answer. It will then proceed down the new path as it had before, backtracking as it encounters dead-ends. Besides, Self-consistancy Chain-of-Thought~\cite{self-consistency-wang} can be expressed in reward form with majority voting as a priority. 

\[
\mathcal{R}_{\text{DFS}}(s_n, a_n^i) = 
\begin{cases} 
1 & \text{if } i = \inf \{j | a_n^j \text{ not visited}\},  \\
0 & \text{otherwise}.
\end{cases}
\]
where $\inf \{j | a_n^j \text{ not visited}\}$ represents the smallest index $j$ among all actions $a_n^j$ that have not been visited.





\subsection{Heuristic Reward}
If only take confirmed priority for one-hot reward, the search process becomes aimless leading to low efficiency. Heuristic search algorithms are designed to solve the problem of search efficiency, such as Greedy Best First Search (GBFS), Dijkstra and A*. Aligned with the characteristic of algorithms, Heuristic reward defined by the heuristic function $h(s)$. Here,we would like to take GBFS for an example and list other heuristic reward in the appendix. the distance from the current state $s_n$ to the target state $s_g$ is used as the heuristic reward, leading the search direction correctly. Given a heuristic function $h(s)$ estimating the cost from any state $s$ to the goal state $s_g$, the heuristic reward for an action $a_n^i$ at state $s_n$ is defined as follows:

\begin{align*}
& \mathcal{R}_{\text{GBFS}}(s_n, a_n^i) &\\
&= \begin{cases} 
h(s_{n+1}) & \text{if } s_{n+1} \text{ is reached by } a_n^i, \\
-\infty & \text{otherwise},
\end{cases}
\end{align*}

\noindent where $h(s_{n+1})$ represents the heuristic cost from the resulting state $s_{n+1}$, after taking action $a_n^i$, to the goal state $s_g$. The action leading to the state with the lowest heuristic cost is preferred, guiding the search towards $s_g$.


\subsection{Simulated rewards}
With the fixed heuristic function for reward, it is evident that most of the decision space lacks coverage, resulting in insufficient exploration for searching. In contrast, simulated search algorithms like MCTS, would explore exhaustively within entire decision space. In this kind of algorithms, an iterative simulation cycle would continue until a terminal state arrived, which usually encompasses three phases: selection, expansion and backpropagation. Alongside the simulation process, a state-action value function $Q(s_n,a_n)$ is maintained, indicating the expected future reward lf taking action $a_n$ in state $s_n$. To control the balance between exploration and exploitation, Upper Confidence bounds applied to Trees is often used. For each iteration of simulation, the selected action  \( a^* \) should be :

\begin{equation*}
a_n^* = \underset{a_n \in A_n}{\mathrm{argmax}} \left[ Q(s_n, a_n) + w \sqrt{\frac{N(s_n)}{1+N(s_n,a_n)}} \right],
\end{equation*}
where \( N(s) \) is the number of times state \( s \) has been visited in previous iterations, \( N(s_n, a_n) \) is the number of times that $a_n$ is selected at the state $s_n$, and weight $w$ controls the proportion of exploration and development.

If taking MCTS as an example and supposed that to abtain the reward of an action needs simulate $d$ times, simulated rewards can be expressed as follow: 

\[
\mathcal{R}_{\text{MCTS}}(s_n, a_n^i) = \frac{1}{N(s_n, a_n^i)} \sum_{k=1}^{N(s_n, a_n^i)} Q(s_n, a_n^k)\,.
\]

\section{Prompt}\label{appendix:prompt}

For transition in SRM, we prompt:

\begin{tcolorbox}[colback=gray!10,colframe=black!90,title=Prompt]
For each sub-question, please answer it in a complete sentence that includes your reasoning. And the last sentence ends with \texttt{"\{answer\_instruction\}"} followed by a concise answer.
\end{tcolorbox}

To apply CoT, we prompt:

\begin{tcolorbox}[colback=gray!10,colframe=black!90,title=Prompt, breakable]
Q: Natalia sold clips to 48 of her friends in April, and then she sold half as many clips in May. How many clips did Natalia sell altogether in April and May? \\
A: Natalia sold 48 clips in April and half as many clips in May, so she sold \( 48 \div 2 = 24 \) clips in May. Altogether, she sold \( 48 + 24 = 72 \) clips. The answer is \textbf{72}. \\

Q: James writes a 3-page letter to 2 different friends twice a week. How many pages does he write a year? \\
A: James writes a 3-page letter to 2 different friends twice a week, so he writes \( 3 \times 2 \times 2 = 12 \) pages every week. There are 52 weeks in a year, so he writes \( 12 \times 52 = 624 \) pages a year. The answer is \textbf{624}. \\

Q: Alexis is applying for a new job and bought a new set of business clothes to wear to the interview. She went to a department store with a budget of \textdollar{}200 and spent \textdollar{}30 on a button-up shirt, \textdollar{}46 on suit pants, \textdollar{}38 on a suit coat, \textdollar{}11 on socks, and \textdollar{}18 on a belt. She also purchased a pair of shoes, but lost the receipt for them. She has \textdollar{}16 left from her budget. How much did Alexis pay for the shoes? \\
A: Alexis spent \textdollar{}30 on a button-up shirt, \textdollar{}46 on suit pants, \textdollar{}38 on a suit coat, \textdollar{}11 on socks, and \textdollar{}18 on a belt, so she spent  
\[
30 + 46 + 38 + 11 + 18 = \textdollar{}143
\]
on everything else. Alexis had a budget of \textdollar{}200 and finally there was \textdollar{}16 left, so she spent  
\[
200 - 16 = \textdollar{}184
\]
in total. Since Alexis has spent \textdollar{}143 on everything else, she spent  
\[
184 - 143 = \textdollar{}41
\]
on the shoes. The answer is \textbf{41}. \\

Q: Ken created a care package to send to his brother, who was away at boarding school. Ken placed a box on a scale, and then he poured into the box enough jelly beans to bring the weight to 2 pounds. Then, he added enough brownies to cause the weight to triple. Next, he added another 2 pounds of jelly beans. And finally, he added enough gummy worms to double the weight once again. What was the final weight of the box of goodies, in pounds? \\
A: Ken poured jelly beans into the box until the weight was 2 pounds, so the weight of the box was 2 pounds at first. Then Ken added enough brownies to cause the weight to triple, so the weight of the box was  
\[
2 \times 3 = 6
\]
pounds. After Ken added another 2 pounds of jelly beans, the weight of the box was  
\[
6 + 2 = 8
\]
pounds. Finally, he added enough gummy worms to double the weight once again, so the weight of the box was  
\[
8 \times 2 = 16
\]
pounds. The answer is \textbf{16}. \\

Q: Janet’s ducks lay 16 eggs per day. She eats three for breakfast every morning and bakes muffins for her friends every day with four. She sells the remainder at the farmers' market daily for \textdollar{}2 per fresh duck egg. How much in dollars does she make every day at the farmers' market? \\

A: Janet’s ducks lay 16 eggs per day. She consumes  
\[
3 + 4 = 7
\]
eggs daily, leaving her with  
\[
16 - 7 = 9
\]
eggs to sell. Since each egg sells for \textdollar{}2, her total daily earnings are  
\[
9 \times 2 = \textdollar{}18.
\]
The answer is \textbf{18}.
\end{tcolorbox}

To get the transited state for the given action and state in BW, we prompt:

\begin{tcolorbox}[colback=gray!10,colframe=black!90,title=Prompt, breakable]
I am playing with a set of blocks where I need to arrange the blocks into stacks. Here are the actions I can do: \\
- Pick up a block \\
- Unstack a block from on top of another block \\
- Put down a block \\
- Stack a block on top of another block \\

\textbf{I have the following restrictions on my actions:} \\
- I can only pick up or unstack one block at a time. \\
- I can only pick up or unstack a block if my hand is empty. \\
- I can only pick up a block if the block is on the table and the block is clear. A block is clear if the block has no other blocks on top of it and if the block is not picked up. \\
- I can only unstack a block from on top of another block if the block I am unstacking was really on top of the other block. \\
- I can only unstack a block from on top of another block if the block I am unstacking is clear. Once I pick up or unstack a block, I am holding the block. \\
- I can only put down a block that I am holding. \\
- I can only stack a block on top of another block if I am holding the block being stacked. \\
- I can only stack a block on top of another block if the block onto which I am stacking the block is clear. Once I put down or stack a block, my hand becomes empty. \\

After being given an initial state and an action, give the new state after performing the action.

\textbf{[SCENARIO 1]} \\  
\textbf{[STATE 0]} \\
I have that, the white block is clear, the cyan block is clear, the brown block is clear, the hand is empty, the white block is on top of the purple block, the purple block is on the table, the cyan block is on the table and the brown block is on the table. \par
\textbf{[ACTION]} Unstack the white block from on top of the purple block. \par
\textbf{[CHANGE]} The hand was empty and is now holding the white block, the white block was on top of the purple block and is now in the hand, the white block is no longer clear, and the purple block is now clear. \par
\textbf{[STATE 1]} \\
I have that, the purple block is clear, the cyan block is clear, the brown block is clear, the hand is holding the white block, the white block is in the hand, the purple block is on the table, the cyan block is on the table and the brown block is on the table. \\

\textbf{[SCENARIO 2]} \\  
\textbf{[STATE 0]} \\
I have that, the purple block is clear, the cyan block is clear, the white block is clear, the hand is empty, the cyan block is on top of the brown block, the purple block is on the table, the white block is on the table and the brown block is on the table. \par
\textbf{[ACTION]} Unstack the cyan block from on top of the brown block. \par
\textbf{[CHANGE]} The hand was empty and is now holding the cyan block, the cyan block was on top of the brown block and is now in the hand, the cyan block is no longer clear, and the brown block is now clear. \par
\textbf{[STATE 1]} \\
I have that, the purple block is clear, the brown block is clear, the cyan block is in the hand, the white block is clear, the hand is holding the cyan block, the purple block is on the table, the white block is on the table and the brown block is on the table. \\

\textbf{[SCENARIO 3]} \\  
\textbf{[STATE 0]} \\
I have that, the red block is clear, the blue block is clear, the hand is empty, the red block is on top of the yellow block, the blue block is on top of the orange block, the orange block is on the table and the yellow block is on the table. \par
\textbf{[ACTION]} Unstack the red block from the yellow block. \par
\textbf{[CHANGE]} The hand was empty and is now holding the red block, the red block was on top of the yellow block and is now in the hand, the red block is no longer clear, and the yellow block is now clear. \par
\textbf{[STATE 1]} \\
I have that, the yellow block is clear, the blue block is clear, the hand is holding the red block, the red block is in the hand, the blue block is on top of the orange block, the orange block is on the table and the yellow block is on the table. \\

\end{tcolorbox}

%% file: tables/task_illust.tex
\begin{table*}[h]\label{tab:task_alignment}
\centering
\caption{\label{tab:task-alignment}Alignment of Three Decision-making Tasks.
GSM8K and FinQA, differ in complexity and domain, but both numerical reasoning tasks with action space defined by $K$ and requiring LLM for action generation and transition. Instead, in Blocksworld, a more complex planning task, an action is composed of one of the 4 verbs (i.e., stack, unstack, put, and pick) and manipulated objects. Thus, the action set for a given state consists of $m$ actions, with $m$ being up to 4,  generated independently of LLM assistance.}
\resizebox*{!}{0.95\columnwidth}{
\begin{tabular}{m{0.15\textwidth}<{\centering}m{0.24\textwidth}<{\centering}m{0.24\textwidth}<{\centering}m{0.24\textwidth}<{\centering}}
\toprule
 & \textbf{GSM8K} & \textbf{FinQA} & \textbf{Blocksworld} \\
\hline
\textbf{Goals} &
Calculate the correct answer by multi-step mathematical reasoning. &
Calculate the correct answer by numerical reasoning for financial problems. &
Arrange the blocks into stacks on a table in the specific order. \\
\hline
\textbf{Initial State $s_0$} &
$\emptyset$ &
$\emptyset$ &
Description of current blocks and a goal. \\
\hline
\textbf{Goal State $s_g$} &
A correct series of problem decomposition leading to the final answer. &
A correct series of problem decomposition leading to the final answer. &
A feasible plan including series actions. \\
\hline
\textbf{State ${s_n}$} &
All current sub-questions and answers. &
All current sub-questions and answers. &
Text description of the current orientation of the blocks. \\
\hline
\textbf{Action Set $A_n$} &
$K$ sub-questions &
$K$ sub-questions &
$m$ actions, $m \le 4$ \\
\bottomrule
\end{tabular}}
\end{table*}

%% file: tables/bw_steps_result.tex
\begin{table}[!t]
\caption{Performance comparison between CoT and MCTS methods, with and without SRM, across different step sizes in Blocksworld (BW) tasks. Results are shown for both Easy and Hard modes, evaluating accuracy at 2-step, 4-step, 6-step, 8-step, 10-step, 12-step, and overall (All) steps.}
\label{tab:comparison-cot-rap-methods_bw}
\centering
\small
\resizebox{\columnwidth}{!}{ 
\begin{tabular}{l l c c c c c c c}
\toprule
\bf Mode & \bf Method & \bf 2-step & \bf 4-step & \bf 6-step & \bf 8-step & \bf 10-step & \bf 12-step & \bf All \\
\midrule
\multirow{4}{*}{Easy}    
& CoT         & 0.49 & 0.18 & 0.06 & 0.01 & 0.01 & 0.00 & 0.08 \\
& MCTS        & 1.00 & 0.99 & 0.75 & 0.61 & 0.32 & 0.32 & 0.66 \\
& \textbf{MCTS } &  &  &  &  &  &  &  \\
& \textbf{\hspace{0.2em} + SRM } & 1.00 & 0.97 & 0.70 & 0.63 & 0.33 & 0.33 & 0.66 \\
& \textbf{MCTS } &  &  &  &  &  &  &  \\
& \textbf{\hspace{0.2em} + SRM$^+$} & 1.00 & 0.99 & 0.76 & 0.65 & 0.33 & 0.35 & 0.68 \\
\midrule
\multirow{4}{*}{Hard}    
& CoT         & 0.22 & 0.14 & 0.02 & 0.02 & 0.00 & 0.00 & 0.05 \\
& MCTS        & 0.67 & 0.76 & 0.74 & 0.48 & 0.17 & 0.09 & 0.51 \\
& \textbf{MCTS   } &  &  &  &  &  &  &  \\
& \textbf{\hspace{0.2em} + SRM } & 0.65 & 0.74 & 0.73 & 0.48 & 0.23 & 0.11 & 0.49 \\
& \textbf{MCTS } &  &  &  &  &  &  &  \\
& \textbf{\hspace{0.2em} + SRM$^+$} & 0.68 & 0.79 & 0.78 & 0.55 & 0.31 & 0.15 & 0.54 \\
\bottomrule
\end{tabular}}
\end{table}

Building on these results, Table~\ref{tab:comparison-cot-rap-methods_bw} provides further evidence of SRM's effectiveness in both \textbf{Easy} and \textbf{Hard} modes of Blocksworld. While MCTS enhances decision-making, SRM maintains similar performance with much lower cost. In \textbf{Hard} mode, SRM$^+$ consistently improves accuracy, especially in complex tasks like the 12-step problems. These findings confirm that SRM reduces cost while preserving performance, and SRM$^+$ further extends this by improving results in more challenging scenarios.